\UseRawInputEncoding
\documentclass[letterpaper, 10 pt, conference]{ieeeconf}  % Comment this line out if you need a4paper

\IEEEoverridecommandlockouts                              % This command is only needed if 
\usepackage{amsfonts}
\usepackage{amsmath}
\usepackage{multirow}
\usepackage{longtable}
\usepackage{tabularx}
\usepackage{lscape}
\usepackage{graphicx}
\usepackage{subcaption}
\graphicspath{ {./Images/} }
\usepackage{xcolor}
\title{\LARGE \bf
SST-Calib: Simultaneous Spatial-Temporal Parameter Calibration between LIDAR and Camera
}

\author{
Akio Kodaira$^{*}$$^{1}$, Yiyang Zhou$^{*}$$^{1}$, Pengwei Zang$^{1}$, Wei Zhan$^{1}$, and Masayoshi Tomizuka$^{1}$ % <-this % stops a space
\thanks{$^{*}$Equal Contribution}
\thanks{$^{1}$Mechanical Systems Control Lab, University of California, Berkeley, CA, USA, 94705}
\thanks{Correspondence: \texttt{akio.kodaira@berkeley.edu}}
}

\begin{document}

\maketitle
\thispagestyle{empty}
\pagestyle{empty}

%%%%%%%%%%%%%%%%%%%%%%%%%%%%%%%%%%%%%%%%%%%%%%%%%%%%%%%%%%%%%%%%%%%%%%%%%%%%%%%%
\begin{abstract}
With information from multiple input modalities, sensor fusion-based algorithms usually out-perform their single-modality counterparts in robotics. Camera and LIDAR, with complementary semantic and depth information, are the typical choices for detection tasks in complicated driving environments. For most camera-LIDAR fusion algorithms, however, the calibration of the sensor suite will greatly impact the performance. More specifically, the detection algorithm usually requires an accurate geometric relationship among multiple sensors as the input, and it is often assumed that the contents from these sensors are captured at the same time. Preparing such sensor suites involves carefully designed calibration rigs and accurate synchronization mechanisms, and the preparation process is usually done offline. In this work, a segmentation-based framework is proposed to jointly estimate the geometrical and temporal parameters in the calibration of a camera-LIDAR suite. A semantic segmentation mask is first applied to both sensor modalities, and the calibration parameters are optimized through pixel-wise bidirectional loss. We specifically incorporated the velocity information from optical flow for temporal parameters. Since supervision is only performed at the segmentation level, no calibration label is needed within the framework. The proposed algorithm is tested on the KITTI dataset, and the result shows an accurate real-time calibration of both geometric and temporal parameters. 
\end{abstract}

%%%%%%%%%%%%%%%%%%%%%%%%%%%%%%%%%%%%%%%%%%%%%%%%%%%%%%%%%%%%%%%%%%%%%%%%%%%%%%%%
\section{INTRODUCTION}
Robots nowadays are usually equipped with multiple types of sensors to ensure a comprehensive perception of the scene. Different sensors on a robot provide observations with their unique physical characteristics. For mobile robots in particular, cameras and LIDARs are among the most popular choices in perception tasks. In outdoor scenes, camera images provide semantically rich representation in dense 2D formats without any depth data, and LIDARs offer accurate yet sparse 3D measurements as point clouds.
    
The complementary nature of these two types of sensors inspires the designs of fusion algorithms for detection tasks. In practice, the fusion algorithms are proved to out-perform single-modal methods with a significant margin \cite{fusionReview}. More importantly, since multiple sensors are located differently on the vehicle, a complete sensor suite can often overcome potential occlusion problems in a single sensor setup. 

One of the fundamental elements in sensor fusion algorithms is calibration \cite{fusionReview}. To begin with, the accurate geometric relationship between two sensor modalities is required for most fusion algorithms \cite{automap1,automap2,ziningfusion,kiwooroar,lls}. More specifically, three translation and three rotation parameters (six degrees of freedom or 6-DOF) need to be provided as inputs. However, acquiring such geometrical relationships is challenging. Human labeling, artificial rigs, and manual measurements are often involved in some earlier attempts, and such processes are usually time-consuming. More detrimentally is the irreproducible nature of the one-time calibration: whenever the vehicle experiences vibration or collision, the geometric data from manual calibration are voided. Thus, automatically calibrating the geometry is critical for any onboard sensor suite. In recent years, a series of learning-based methods have started to predict the extrinsic parameters of the sensor suite. These methods are usually data hunger, requiring pre-labeled data as supervision to predict the parameters in the target sensor setting \cite{RegNetMS, calibnet, LCCNet}. However, the ground truth calibration data is still hard to obtain and none of these works reported the generalization capacity on other datasets. Some other methods leverage the semantic outcome of each sensor modality and directly regress the extrinsic matrices without supervision. This work follows on SemAlign \cite{liu2021semalign} for finer geometric calibration. 

Beyond the geometric parameter estimation, the temporal relationship among sensors also requires researchers' attention. More specifically, how to synchronize each sensor remains to be a long-neglected problem in the autonomous driving research domain. Algorithms usually assume naively that the content in each sensor frame is acquired at precisely the same time, but this assumption rarely holds. For example, in the Argoverse dataset \cite{Argoverse}, the synchronization is performed by simply associating the closest timestamp between LIDARs and cameras. Some other works \cite{urbanloco} perform synchronization at the data acquisition level, but the static nature of a one-time delay compensation could not address the changes in the real-world environment. Thus, dynamically calibrating the time delay among sensors is critical, yet almost unexplored.

This work proposes a joint spatial-temporal calibration framework between LIDARs and cameras on an autonomous driving platform. The input of the proposed framework are sequences of camera and LIDAR frames. Here, each sensor modality is processed through an arbitrary semantic segmentation network, which one can choose based on the available training data. Secondly, the segmented LIDAR point cloud is projected onto the semantic image, where a newly designed bi-directional alignment loss is calculated for geometrical parameter regression. Not limited to point-to-pixel loss, we down-sampled the semantic pixel for pixel-to-point loss as well. To estimate the time delay between sensor modalities, we estimate the visual odometry from two consecutive images and predict a shifted point cloud for matching.

The joint spatial-temporal calibration framework is tested on the KITTI dataset for 150 frames, and the proposed method can robustly regress geometrical and temporal relationships among LIDARs and cameras. Ablation studies are also added to demonstrate the bi-directional loss superiority and the strong temporal estimation capacity. 

The major contributions of this work are:

\begin{itemize}
    \item A joint spatial-temporal calibration algorithm is proposed for LIDAR-camera sensor suite;
    \item A bi-directional loss was designed for more robust performance in geometrical parameter regression;
    \item A time parameter was coupled with visual odometry to estimate the temporal delay among sensors.
\end{itemize}

\section{Related work}
In this section, we will first review independent geometric and temporal calibration works for LIDAR-camera suite. A summary of the joint calibration works will be included at the end.

\subsection{Geometric Calibration}

\subsubsection{Rig-based Calibration}
The earliest attempts in geometric sensor calibration start with artificial targets, also known as rigs. Such objects are often of regular shapes or specific patterns, making them easy for simple detection algorithms.
            
In Single-Shot Calibration, Geiger et al. propose to use checkerboards as targets for the calibration task between cameras and range sensors \cite{SSC}. The patterns on the checkerboard are easy to identify for cameras, and the board themselves are easily detectable planes for range sensors. A downstream optimization process is then carried out for the alignment. Following a similar pipeline, other methods use different rigs or features for such calibration tasks \cite{kimboard,sphericalCalib,edgecalib}. However, constructing an artificial rig could consume a significant amount of time, and such a one-time calibration result is often not repeatable. For example, if the sensor suite is changed geometrically in accidents, a new round of rig calibration is required. 

\subsubsection{Feature-based Calibration}
Departing from the calibration board, some other methods explore the possibilities of finding various features in the real world for calibration. 
            
Exploring the geometric features in the environment inspires the authors of \cite{edgeindoor} and \cite{lineindoor} to use edges and lines in indoor scenes to estimate the 6-DOF of a sensor suite. Tamas et al., on the other hand, use the plane feature in outdoor environments \cite{planeoutdoor}. However, these features are not universally available in all scenes, and advanced detection/correspondence algorithms are often needed. Another school of thought is to utilize the color channels in images and intensity channels in LIDARs to maximize the mutual information for calibration purposes \cite{Pandey2015AutomaticEC}. However, depending on the viewing angle, the intensity return from the LIDAR could be drastically different from the image outcomes \cite{glassLIDAR}. 

\subsubsection{End-to-end Calibration}
More recently, some learning-based methods predict the extrinsic parameters of the sensor suite with raw sensor inputs \cite{RegNetMS, calibnet, LCCNet}. To train these prediction models, researchers usually need to acquire a large dataset with ground-truth calibration data. More importantly, none of the aforementioned works evaluate the models' generalization capacity on other datasets.
            
Last year, SemAlign was proposed as a semantic segmentation-based extrinsic calibration method \cite{liu2021semalign}. The model utilizes off-the-shelf segmentation networks for LIDARs and cameras, and it optimizes the geometric calibration parameters for minimum Semantic Alignment loss. Such a method is favorable because the generalization capacity lies within the semantic segmentation modules, which are well explored domains in computer vision. However, the single-direction loss function in this work is not robust enough for complicated scenes.  

\subsection{Temporal Calibration}
Sensor synchronization is often overlooked in driving-related tasks \cite{fusionReview}, and most algorithms often naively assume that the data provided is already synchronized. However, such assumptions rarely hold. To synchronize the sensors, researchers designed both offline and online mechanisms for synchronization. Here we introduce a few existing attempts related to temporal calibration. 

\subsubsection{Offline Synchronization}
In offline designs, the synchronization is performed before the data acquisition. Similar to the triggering mechanism in multi-cameras setups in the early days of photography, a naive attempt in sensor synchronization is to use hardware triggers \cite{Chen2020AdvancedMR} like a fix-frequency impulse signal. Some other datasets prepare software triggers with compensated triggering time \cite{urbanloco}, but the manual calibration of the time delay is usually time-consuming. More recently, the IEEE1588 Precision Time Protocol is also introduced in system-wide synchronization applications.

\subsubsection{Online Synchronization}
Online synchronization happens during or after the data acquisition process. A typical example is the Argoverse dataset\cite{Argoverse}, where each camera image is matched to the closest LIDAR scan, while the physical sensor themselves are not exactly synchronized. Most research work in this domain estimates the time delay between sensors while they capture real-world data. Visual-Inertial Navigation Systems (VINS) calibrates the time delay between the high-frequency Inertia Measurement Unit (IMU) and a camera by optimizing the trajectory consistency between these two sensors.

\subsection{Joint Calibration}
Calibration through Simultaneously Localization and Mapping (SLAM) is a popular method for the joint calibration task. Through SLAM algorithms, each sensor modality provides unique odometry that is both spatial and temporal dependent. In \cite{Park2020SpatiotemporalCC} and \cite{7555301}, the authors utilize such odometry estimation to find the best spatial-temporal match between the two generated trajectories. However, the aforementioned methods leverage the performance of the SLAM algorithms over the whole traveling trajectory, and SLAM's drifting effect could easily deteriorate the performance of the downstream calibration task. In our proposed work, we optimize the extrinsic and the temporal parameter iteratively, avoiding error accumulation along the traveling path. Furthermore, Persic et al. propose to use tracking results from both sensor modalities to solve for geometric and temporal calibration parameters \cite{9387269}. Similarly, the calibration performance is also limited by the accumulated error in tracked objects.  

\begin{figure*}[!t]
    \centering
    \includegraphics[width=170mm]{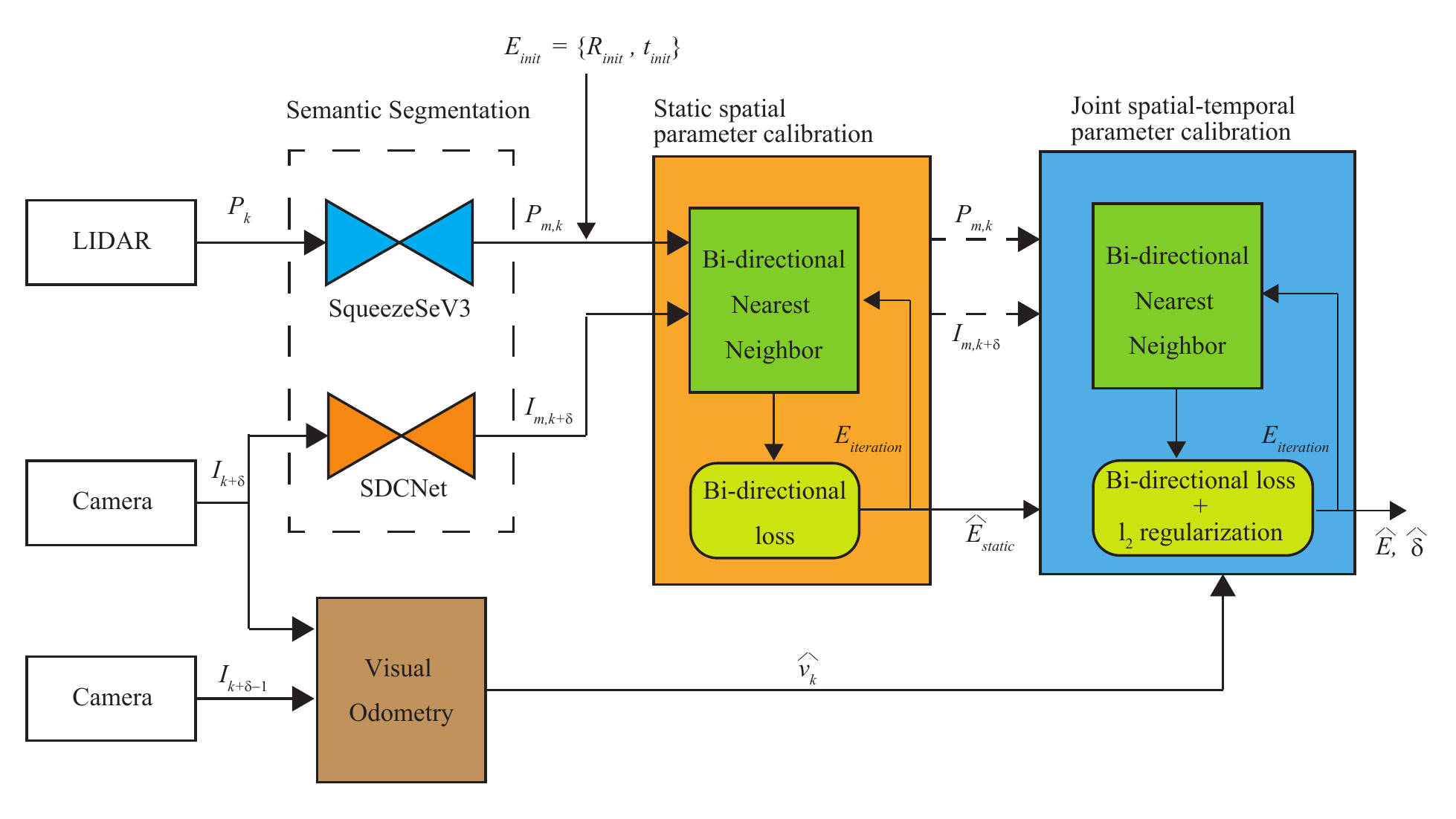}
    \caption{Workflow of the proposed calibration method.}
    \label{fig:work_flow}
\end{figure*}

\section{Methodology}
    The workflow of the proposed calibration method is shown in Figure \ref{fig:work_flow}. The calibration process consists of the static spatial parameter calibration module for the spatial initial guess and the joint spatial-temporal parameter calibration module for duo-parameter estimation. 
    
    The inputs of the proposed algorithm are one point cloud scan $P_k \in \mathbb{R}^{3\times N_p}$, and two sequential RGB images $\{I_{k+\delta}, I_{k+\delta - 1}\} \in \mathbb{Z}^{N_h \times N_w \times 3}$. Where $N_p$ is the number of the points in the scan, and $N_h$ and $N_w$ are the image dimensions. The goal of the algorithm is to estimate a 6-DOF $\{R, \textbf{\textit{t}}\}$ ($R\in \mathbb{R}^{3\times3}, t\in \mathbb{R}^{3}$) of the geometric relationship and the time delay $\delta \in \mathbb{R}$ between $P_k$ and $I_{k+\delta}$.
    
    To do this, we first process $P_k$ and $I_{k+\delta}$ through arbitrary semantic segmentation algorithms to obtain semantic mask $P_{m,k}$, $I_{m,k+\delta}$. Then with an initial extrinsic guess $\{R_{init}, \textbf{\textit{t}}_{init}\}$ from rough measurement or sampling and the known intrinsic $K \in \mathbb{R}^{3\times3}$, we project the LIDAR point cloud onto the camera image plane. By finding the nearest neighbor both from point to pixel and from pixel to point, we calculate the Euclidean distance between them which is the loss function of the optimization algorithm. 
    
    The first optimization iteration (the static spatial parameter calibration module) will be carried out at the frame when the ego vehicle's velocity is almost 0. The static spatial parameter calibration gives the initial estimation of the rotation and translation $\{\hat{R}_{static}, \hat{\textbf{\textit{t}}}_{static}\}$. This estimation will be used as the initial guess and the regularization reference for the joint spatial-temporal parameter calibration.
    
    Secondly, for dynamic scenes, we estimate the temporal information between $I_{k+\delta}$ and $I_{k+\delta - 1}$ from the visual odometry which will predict the velocity $\hat{\textbf{\textit{v}}}_k \in \mathbb{R}^{3}$ between two camera frames. Here, the translation shift between $P_k$ and $I_{k+\delta}$ can be represented as $\textbf{\textit{t}}_{\delta, k} = \hat{\textbf{\textit{v}}}_k \cdot \delta$.
    We use this $\hat{\textbf{\textit{v}}}_k$ as part of the optimization and estimate both $\hat{\delta}$, and $\{\hat{R}, \hat{\textbf{\textit{t}}}\}$.
    
    \subsection{Semantic Segmentation}
    With off-the-shelf semantic segmentation modules, the proposed method would generalize to any dataset with semantic labels. In this paper, we use SqueezeSegV3\cite{xu2020squeezesegv3} and SDC-net\cite{reda2018sdc} for point cloud and image semantic segmentation, respectively. Considering the frequent presence of vehicles in the urban environment, in this work, we only use the car class for semantic segmentation. Applying these semantic segmentation modules to input, we get semantic masks $P_{m,k}$, $I_{m,k+\delta}$.
    
    \subsection{Point Cloud Projection}
    In order to compute the semantic loss, we first project the semantic mask of a point $p_{i,m,k} \in P_{m,k}$ ($p_{i,m,k}\in \mathbb{R}^{3}$) onto the 2D image plane. Based on the classic camera model \cite{invitationTo3dVision}, we can achieve the projection as follows,
    
    \begin{equation}
        \begin{bmatrix}
        px_{i,m,k}\\
        py_{i,m,k} \\
        pz_{i,m,k} \\
        \end{bmatrix}
        = K[R\textbf{\textit{p}}_{i,m,k} + \textbf{\textit{t}}]
    \end{equation}
    
    \begin{equation}
        \\
        \Bar{\textbf{\textit{p}}}_{i,m,k} = \\
        \\
        \begin{bmatrix}
        pu_{i,m,k}\\
        \\
        pv_{i,m,k} \\
        \end{bmatrix}
        = \frac{1}{pz_{i,m,k}}
        \begin{bmatrix}
        px_{i,m,k}\\
        \\
        py_{i,m,k} \\
        \end{bmatrix}
    \end{equation}
    
    where, $pu_{i,m,k}$ and $pv_{i,m,k}$ are the image coordinates of the projected point $\Bar{\textbf{\textit{p}}}_{i,m,k} \in \mathbb{R}^{2}$.

    \subsection{Bi-directional loss}
    Let ${\Bar{\textbf{\textit{p}}}_{1,m,k}...\Bar{\textbf{\textit{p}}}_{n_p,m,k}}$ be a set of projected LIDAR points that are within the camera's field of view. Now for the projected point $\Bar{\textbf{\textit{p}}}_{i,m,k}$, let $\textbf{\textit{q}}_{j,m,k+\delta} \in I_{m, k+\delta}$ be the nearest neighbor pixel of the same class. Then, the single-direction point-to-pixel (point-to-image) semantic alignment loss on frame $k$ can be computed as follows,
    
    \begin{equation}
        L_{p2i,k} = \sum_i||\Bar{\textbf{\textit{p}}}_{i,m,k}(R,t,\textbf{\textit{p}}_{i,m,k})- \textbf{\textit{q}}_{j,m,k+\delta}||^2_2
    \end{equation}

\begin{figure}
\begin{subfigure}{.25\textwidth}
  \centering
  \includegraphics[width=.8\linewidth]{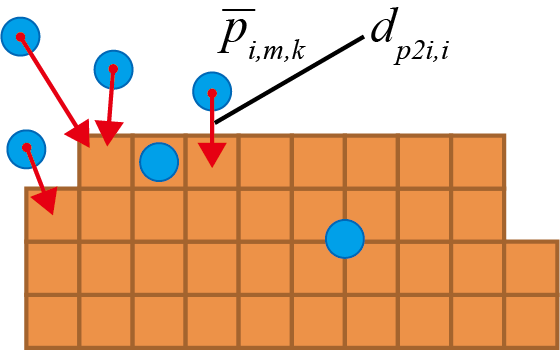}
  \caption{Point-cloud to Pixels}
  \label{fig:sfig1}
\end{subfigure}%
\begin{subfigure}{.25\textwidth}
  \centering
  \includegraphics[width=.8\linewidth]{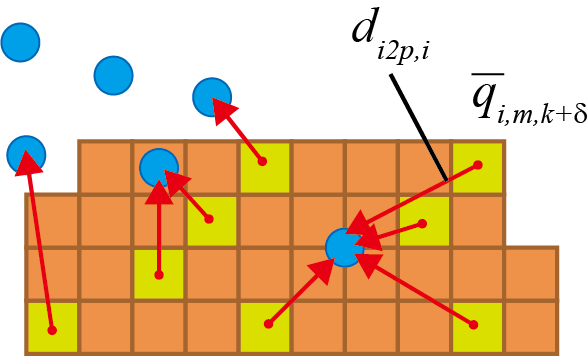}
  \caption{Pixels to Point-cloud}
  \label{fig:sfig2}
\end{subfigure}
\caption{Bi-directional Projection Demonstration: Here, blue circles correspond to projected points, and orange squares represent image pixels. Yellow squares highlight the down-sampled pixels.}
\label{fig:fig_bi}
\end{figure}

    Here, the loss is computed per projected point. Figure \ref{fig:sfig1} demonstrates the point-to-pixel loss calculation process. As shown in \cite{liu2021semalign}, by minimizing this loss function, we can make the projected point cloud well overlapped with the pixels that have the same semantic label. Thus, minimizing this loss function could lead us to the correct estimation for $\hat{E}_{static}=\{ \hat{R}_{static},\hat{\textbf{\textit{t}}}_{static}\}$.
    
    However, when the initial guess for the extrinsic matrix is significantly different from the ground truth, the nearest neighbor matching does not necessarily give us the appropriately matching result for most of the pairs, and the information of some important pixels will be abandoned. Thus, minimizing the single-directional loss would fall into the inappropriate local minimum.
    
    To avoid such loss of information, we propose a bi-directional loss that utilizes the pixel-to-point (image-to-point) nearest neighbor matching as well (Figure \ref{fig:sfig2}).
    
    Considering the fact that one image has too many pixels to match in real-time, we down-sample the pixels for the pixel-to-point matching. Let $\{\Bar{\textbf{\textit{q}}}_{1,m,k+\delta}...\Bar{\textbf{\textit{q}}}_{n_i,m,k+\delta}\} \subset I_{m, k+\delta} $ be a set of the down-sampled pixels. Now for the pixel $\Bar{\textbf{\textit{q}}}_{i,m,k+\delta}$,  $\Bar{\textbf{\textit{p}}}_{j,m,k} \in P_{m,k}$ is the nearest neighbor projected point. Then, the pixel-to-point semantic alignment loss on frame $k$ can be computed as follows,
    
    \begin{equation}
        L_{i2p,k} = \sum_i||\Bar{\textbf{\textit{q}}}_{i,m,k+\delta}- \Bar{\textbf{\textit{p}}}_{j,m,k}(R,t)||^2_2
    \end{equation}
    
    Here, the loss is computed per sampled pixel. Then, the bi-directional semantic alignment loss at the $l$-th iteration can be represented as follows,
    
    \begin{equation}
        L_{bi,k,l} = L_{p2i,k,l} + w_l \cdot \frac{n_p}{n_i} \cdot L_{i2p,k,l}
    \end{equation}
    
    where, $\frac{n_p}{n_i}$ is the normalization term and $w_l \in \mathbb{R}$ is the weight at the optimization iteration number $l$. With smaller $w_l$, the optimizer tends to project the projected points within the image mask to minimize $L_{p2i,k,l}$. With larger $w_l$ for minimizing $L_{i2p,k,l}$, the optimizer tends to have the image mask well included in the projected point cluster. Thus, shifting the value of $w_l$ during optimization iterations can avoid getting stuck in local minimums, and the optimization solution would lead us to the better nearest neighbor matching for the next iteration. Thus, optimizing the bi-directional loss function alone would yield a more refined guess $\hat{E}_{static}=\{ \hat{R}_{static},\hat{\textbf{\textit{t}}}_{static}\}$ for the joint calibration.

\subsection{Joint Spatial-Temporal calibration}
Before the joint calibration, we extract the velocity $\hat{\textbf{\textit{v}}}_k$ between two sequential RGB images $\{I_{k+\delta}, I_{k+\delta - 1}\}$ using visual odometry. The visual odometry used in this paper is based on the sparse optical \cite{lucas1981iterative} flow for FAST feature tracking, and the Nister's 5-point algorithm with RANSAC \cite{nister2004efficient} for essential matrix estimation.

With a moving ego vehicle and an asynchronized sensor suite, the projected point cloud obtained from Eq(2) will never match with the corresponding pixels, even with the ground truth geometric calibration parameters. To compensate for this time delay, we need to modify the projection equation as follows,

\begin{equation}
    \begin{bmatrix}
    px_{i,m,k,\delta}\\
    py_{i,m,k,\delta} \\
    pz_{i,m,k,\delta} \\
    \end{bmatrix}
    = K[\hat{R}_{static}\textbf{\textit{p}}_{i,m,k} + \hat{\textbf{\textit{t}}}_{static} + \hat{\textbf{\textit{v}}}_k \cdot \hat{\delta}]
\end{equation}

\begin{equation}
    \\
    \Bar{\textbf{\textit{p}}}_{i,m,k,\hat{\delta}} = \\
    \\
    \begin{bmatrix}
    pu_{i,m,k,\hat{\delta}}\\
    \\
    pv_{i,m,k,\hat{\delta}} \\
    \end{bmatrix}
    = \frac{1}{pz_{i,m,k,\hat{\delta}}}
    \begin{bmatrix}
    px_{i,m,k,\hat{\delta}}\\
    \\
    py_{i,m,k,\hat{\delta}} \\
    \end{bmatrix}
\end{equation}

$pu_{i,m,k,\hat{\delta}}$ and $pv_{i,m,k,\hat{\delta}}$ are the image coordinate of the projected point compensated with $\hat{\delta}$ and $\hat{\textbf{\textit{v}}}_k$. Therefore, we can estimate both spatial and temporal parameters by minimizing the modified bi-directional loss at iteration $l$.

    \begin{equation}
        L_{bi,k,l,\hat{\delta}} = L_{p2i,k,l,\hat{\delta}} + w_l \cdot \frac{n_p}{n_i} \cdot L_{i2p,k,l,\hat{\delta}} + \beta
    \end{equation}
    
    \begin{equation}
       \beta = \lambda_1||\hat{t} - \hat{t}_{static}||^2_2 +  \lambda_2||\hat{R}\hat{R}_{static}^{-1}||^2_2
    \end{equation}

Here, $\beta$ is the regularization term bringing the estimation closer to initial guesses. $\lambda_1$ and $\lambda_2$ are regularization coefficients for translation and rotation, respectively.

\begin{table*}[!t]
\caption{Joint spatial-temporal calibration for LIDAR and Camera.}
\label{tab:joint}
\resizebox{\textwidth}{!}{%
\begin{tabular}{lcclccccc}
\hline
 &
  \multirow{2}{*}{Time delay {[}ms{]}} &
  \multicolumn{1}{c}{Time delay estimation} &
  \multicolumn{2}{c}{ATD {[}cm{]}} &
  \multicolumn{2}{c}{QAD {[}deg.{]}} &
  \multicolumn{2}{c}{AEAD {[}deg.{]}} \\ \cline{3-9} 
   &
   &
  Average Error {[}ms{]} &
  Mean &
  \multicolumn{1}{l}{Median} &
  Mean &
  Median &
  Mean &
  \multicolumn{1}{l}{Median} \\ \hline

\multicolumn{1}{c}{Taylor et al. \cite{7555301}} &
  100 &
  10.0 &
  \multicolumn{1}{c}{5.27} &
  - &
  - &
  - &
  0.34 &
  - \\ 
\multicolumn{1}{c}{Park et al. \cite{Park2020SpatiotemporalCC}} &
  10 &
  3.5 &
  \multicolumn{1}{c}{20.00} &
  - &
  - &
  - &
  0.88 &
  - \\ \hline

\multicolumn{1}{c}{SSTCalib (ours)} &
  100 &
  3.4 &
  \multicolumn{1}{c}{7.45} &
  5.52 &
  0.67 &
  0.66 &
  0.32 &
  0.32 \\ \hline
\end{tabular}%
}
\end{table*}

\begin{figure*}[!h]
    \centering
    \includegraphics[scale=1]{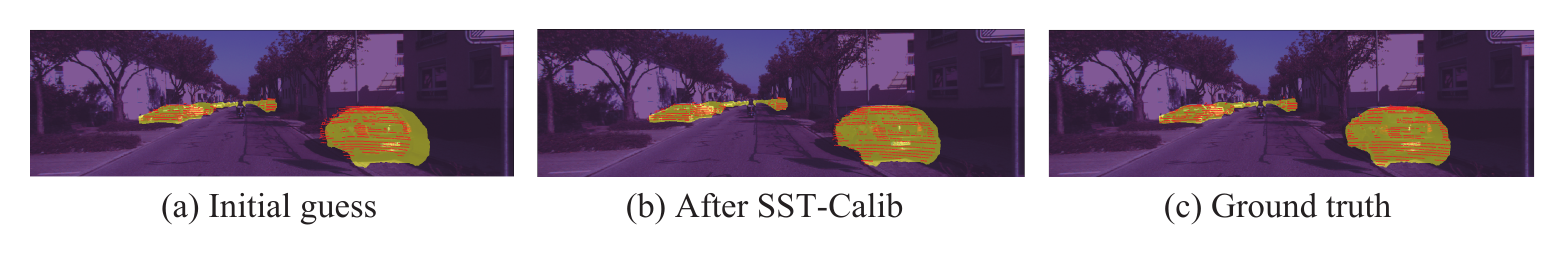}
    \caption{Calibration results on asynchronized KITTI odometry dataset: Red points represent projected point clouds with the car label, the yellow region denotes the pixels with the car label, and the purple region is for the pixels with the other semantic label.}
    \label{fig:my_label}
\end{figure*}

\section{Experiments}

In this section, we report our experiment with the proposed algorithm and related works. After introducing the implementation details, we report our algorithm performance on the joint LIDAR-camera calibration. Furthermore, we show the experiment result of each calibration module under various noise levels in both geometrical and temporal senses. In the end, we include an ablation study on the loss formulation.

\subsection{Implementation Details}
Each driving sequence in KITTI \cite{kitti} includes RGB images, LIDAR point clouds, and accurate extrinsic/intrinsic calibration parameters. We treat intrinsic calibration parameters as given and assume the provided extrinsic parameters as the ground truth. We use Sequence 00 (4541 frames) for evaluation on both static and dynamic calibration.
        
For static calibration, we use $w_l = 20$ for the first 20 optimization iterations, $w_l = 1$, for the next 30 iterations, and $w_l = 0.02$ for the last 10 iterations. The iterative optimization has 60 iterations in total. For the joint calibration part, we use a constant weight ratio $w_l = 5$ for 20 iterations in total. The regularization coefficients we used are $\lambda_1 = 10^6$ and $\lambda_2 = 10^9$. The down-sampling rate of pixels for the pixel-to-point matching is 2\%.
        
We use quaternion angle difference (QAD) and the average Euler angle difference (AEAD) \cite{liu2021semalign} to evaluate the rotational error between estimated rotations and the ground truth rotation. Both angle differences are computed as follows,

\begin{equation}
    QAD = 2 \arccos{(|p\cdot q|)}
\end{equation}

\begin{equation}
    AEAD = (R_{error} + P_{error} + Y_{error})/3
\end{equation}
where $p$ and $q$ are ground truth rotation and estimated rotation, respectively. $R_{error}$, $P_{error}$, $Y_{error}$, denote the absolute Euler angle difference of roll, pitch, and yaw.

ATD (Average Translation Difference) is used to evaluate the absolute translation error. The ATD is calculated as follows,
\begin{equation}
    ATD = (x_{error} + y_{error} + z_{error})/3
\end{equation}
where $x_{error}$, $y_{error}$, $z_{error}$, represent the absolute translation error of x, y, and z axis.

\subsection{Joint Spatial-Temporal Calibration}
In this experiment, it is assumed that the LIDAR and the camera are not synchronized, meaning that there is a certain amount of time delay for each sensor's data acquisition trigger. To exacerbate this delay using the KITTI odometry dataset, we intentionally use RGB image data acquired one frame ahead of the LIDAR point cloud. In the KITTI odometry dataset, both the RGB image and the point cloud are captured every 100~ms (i.e. The data acquisition frequency of both the camera and the LIDAR is 10~Hz.) Therefore, the ground truth of the time delay in this simulation is 100~ms. The initial guesses of translation for each x, y, z-axis are sampled from the uniform distribution [-10~cm, 10~cm]. And the rotation guesses for roll, pitch, and yaw are sampled from a uniform distribution [-10~deg., 10~deg.]. 

\begin{table*}[!t]
\centering
\caption{The static spatial calibration results.}
\label{tab:static}
\resizebox{\textwidth}{!}{%
\begin{tabular}{ccccccccc}
\hline
\multicolumn{1}{l}{} &
  \multirow{2}{*}{Required pre-process} &
  \multicolumn{1}{c}{\multirow{2}{*}{Miss Calibrated Rotation}} &
  \multicolumn{2}{c}{ATD {[}cm{]}} &
  \multicolumn{2}{c}{QAD {[}deg.{]}} &
  \multicolumn{2}{c}{AEAD {[}deg.{]}} \\ \cline{4-9} 
 &
   &
  \multicolumn{1}{c}{} &
  \multicolumn{1}{c}{Mean} &
  \multicolumn{1}{c}{Median} &
  Mean &
  Median &
  Mean &
  \multicolumn{1}{c}{Median} \\ \hline

SemAlign\cite{liu2021semalign} &
  \begin{tabular}[c]{@{}c@{}}Sampling 5000 \\ $E_{init}$ for initialization\end{tabular} &
  \begin{tabular}[c]{@{}c@{}}{[}-10, 10{]}\\ {[}-20, 20{]}\end{tabular} &
  \begin{tabular}[c]{@{}c@{}}-\\ -\end{tabular} &
  \begin{tabular}[c]{@{}c@{}}-\\ -\end{tabular} &
  \begin{tabular}[c]{@{}c@{}}1.14\\ 2.59\end{tabular} &
  \begin{tabular}[c]{@{}c@{}}0.46\\ 0.49\end{tabular} &
  \begin{tabular}[c]{@{}c@{}}0.62\\ 1.49\end{tabular} &
  \begin{tabular}[c]{@{}c@{}}0.23\\ 0.24\end{tabular} \\ \hline
SSTCalib (Ours) &
  \ None &
  \begin{tabular}[c]{@{}c@{}}{[}-10, 10{]}\\ {[}-20, 20{]}\end{tabular} &
  \begin{tabular}[c]{@{}c@{}}18.9\\ 20.2\end{tabular} &
  \begin{tabular}[c]{@{}c@{}}12.8\\ 20.0\end{tabular} &
  \begin{tabular}[c]{@{}c@{}}1.28\\ \textbf{1.53}\end{tabular} &
  \begin{tabular}[c]{@{}c@{}}0.81\\ 1.19\end{tabular} &
  \textbf{\begin{tabular}[c]{@{}c@{}}0.60\\ 0.69\end{tabular}} &
  \begin{tabular}[c]{@{}c@{}}0.38\\ 0.59\end{tabular} \\ \hline
\end{tabular}%
}
\end{table*}

\begin{table*}[!t]
    \caption{Comparison among several time delay}
    \label{tab:time_delay}
    \resizebox{\textwidth}{!}{%
    \begin{tabular}{cccccccc}
    \hline
    \multirow{2}{*}{Time delay {[}ms{]}} &
      \multicolumn{1}{c}{Time delay estimation} &
      \multicolumn{2}{c}{ATD {[}cm{]}} &
      \multicolumn{2}{c}{QAD {[}deg.{]}} &
      \multicolumn{2}{c}{AEAD {[}deg.{]}} \\ \cline{2-8} 
     &
      Average Error {[}ms{]} &
      \multicolumn{1}{c}{Mean} &
      \multicolumn{1}{c}{Median} &
      Mean &
      Median &
      Mean &
      \multicolumn{1}{l}{Median} \\ \hline
    100 & 3.4 & 7.4 & 5.5 & 0.67 & 0.66 & 0.32 & 0.32 \\ \hline
    200 & 13.5 & 6.5 & 6.8 & 0.68 & 0.69 & 0.34 & 0.34 \\ \hline
    300 & 23.3 & 4.6  &  4.9  & 0.70 & 0.67 & 0.34 & 0.33    \\ \hline
    \end{tabular}%
    }
    \label{tb:C2_noisy_time_delay_result}
\end{table*}

\begin{table*}[!t]
\caption{Comparison between single- and bi-direction loss, an ablation study }
\label{tab:robust}
\resizebox{\textwidth}{!}{%
\begin{tabular}{cclcccccc}
\hline
\multicolumn{1}{l}{} &
  \multirow{2}{*}{Initial rotation} &
  \multicolumn{2}{c}{ATD {[}cm{]}} &
  \multicolumn{2}{c}{QAD {[}deg.{]}} &
  \multicolumn{2}{c}{AEAD {[}deg.{]}} &
  \multirow{2}{*}{\begin{tabular}[c]{@{}c@{}}Optimization \\ failure rate {[}\%{]}\end{tabular}} \\ \cline{3-8}
 &
   &
  Mean &
  \multicolumn{1}{l}{Median} &
  Mean &
  Median &
  Mean &
  \multicolumn{1}{l}{Median} &
   \\ \hline
Single-direction loss &
  {[}-10, 10{]} &
  \multicolumn{1}{c}{42.7} &
  22.4 &
  3.67 &
  1.62 &
  1.63 &
  0.79 &
  16.2 \\ \hline
Bi-direction loss &
  {[}-10, 10{]} &
  \multicolumn{1}{c}{\textbf{18.9}} &
  \textbf{12.8} &
  \textbf{1.28} &
  \textbf{0.81} &
  \textbf{0.60} &
  \textbf{0.38} &
  \textbf{8.8} \\ \hline
\end{tabular}%
}
\end{table*}

First, we use Frame 547 to 552 where the car is not moving for the static calibration part. The initial guesses of the spatial parameters are the ground truth with uniform noise. The estimated spatial calibration result is then fed to the joint calibration algorithm to estimate the temporal parameter.
        
We sample 50 frames for joint calibration evaluation. A visualization of the experimental results could be seen in Figure 3. The quantitative results are shown in Table I. Our proposed method estimates the time delay as 103.4~ms, only 3.4~ms different from the ground truth. While appropriately estimating the temporal parameters, the spatial parameters estimation also achieves the same level of performance compared with the state-of-the-art spatial calibration methods.

\subsection{Robustness Study: Calibration under Noises}
 The robustness of our proposed method using the bi-directional loss is demonstrated in this section. We sample 150 frames from sequence 00 with random translation and rotation initial guess error. The initial guess noise for the translation is uniformly distributed within [-10~cm, 10~cm], and the initial guess noise for the rotation is uniformly distributed within [-10~deg., 10~deg.] and [-20~deg., 20~deg.]. In Table II, we observe that our proposed method has a similar level of spatial parameters estimation capability without any costly pre-process. For both rotational noise distribution, our method reports better AEAD results compared to the SemAlign, which samples 5000 random transformations that may lead to better initial guesses. More importantly, when the noise level doubles, our proposed algorithm maintains a similar level of performance. 
        
We also test the algorithm performance under different time delay intervals. Table III shows the result of our experiments with time intervals of 100, 200, and 300 milliseconds. The proposed algorithm shows a successful estimation of both geometric and temporal parameters.

\subsection{Ablation Study: Bi-directional Loss}
As a contribution of the proposed algorithm, we argue that the bi-directional semantic loss is superior to the single-directional loss. The semantic calibration algorithm using single-direction loss is similar to SemAlign \cite{liu2021semalign}, omitting the random pre-sampling process. We report the calibration result of these two loss functions in Table IV. The visualization result is shown in Figure 4. Here, all error metrics of the single-directional loss are approximately twice as large as the ones of the bi-direction loss. More importantly, the single-directional loss formulation fails more frequently at the optimization phase, meaning that the point-to-pixel loss alone is not robust for complicated scenes.

\begin{figure}[!h]
    \centering
    \includegraphics[scale=1]{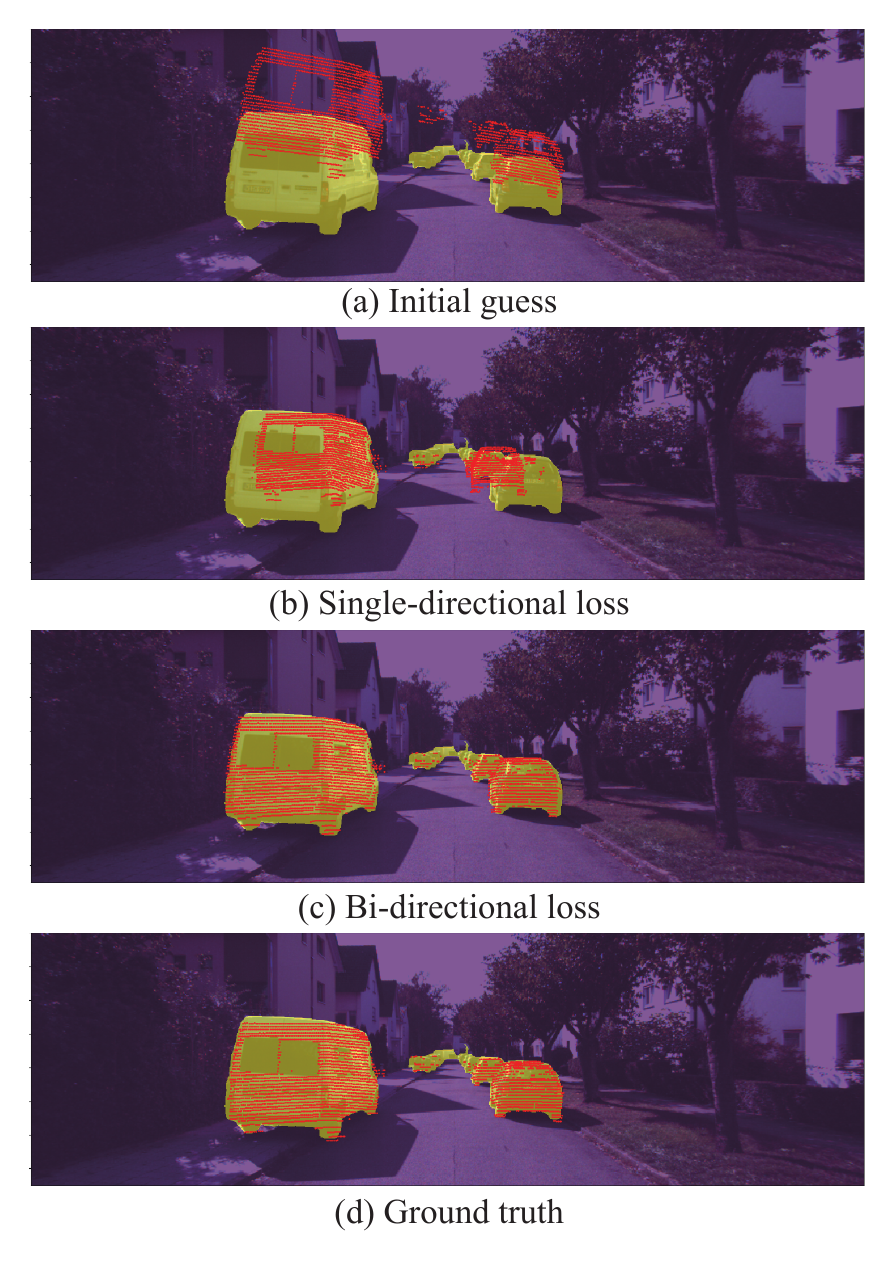}
    \caption{Calibration results of single-directional and bi-directional losses: For a fair comparison, both single-directional and bi-directional losses use the same initial guess (a). The algorithm that only uses the single-directional loss falls into the inappropriate local minimum (b). The algorithm that uses the bi-directional loss obtained a result almost identical to the ground truth (c), (d).}
    \label{fig:ablation}
\end{figure}

\section{Conclusions}
In this paper, we propose an online joint spatial-temporal calibration algorithm for the LIDAR-camera sensor suite in a robotic setup. We design a bi-directional semantic loss in the image domain for a more accurate and robust estimation of the parameters. More importantly, we incorporated the time delay estimation from visual odometry to synchronize different sensor modalities in an online fashion. The proposed algorithm is tested on the KITTI dataset, and various experiments and ablation studies have demonstrated the effectiveness and real-time capability of the proposal. For the next step, we plan to experiment with optical flow strategies for on-image dynamic prediction.


%%%%%%%%%%%%%%%%%%%%%%%%%%%%%%%%%%%%%%%%%%%%%%%%%%%%%%%%%%%%%%%%%%%%%%%%%%%%%%%%

%%%%%%%%%%%%%%%%%%%%%%%%%%%%%%%%%%%%%%%%%%%%%%%%%%%%%%%%%%%%%%%%%%%%%%%%%%%%%%%%

%%%%%%%%%%%%%%%%%%%%%%%%%%%%%%%%%%%%%%%%%%%%%%%%%%%%%%%%%%%%%%%%%%%%%%%%%%%%%%%%
%\section*{APPENDIX}

%\section*{ACKNOWLEDGMENT}
%The preferred spelling of the word ÒacknowledgmentÓ in America is without an ÒeÓ after the ÒgÓ. Avoid the stilted expression, ÒOne of us (R. B. G.) thanks . . .Ó  Instead, try ÒR. B. G. thanksÓ. Put sponsor acknowledgments in the unnumbered footnote on the first page.

%%%%%%%%%%%%%%%%%%%%%%%%%%%%%%%%%%%%%%%%%%%%%%%%%%%%%%%%%%%%%%%%%%%%%%%%%%%%%%%%

\bibliography{ITSC2022}

\end{document}